\documentclass[conference]{IEEEtran}
\usepackage{amsmath,amssymb,amsfonts}
\usepackage{algorithmic}
\usepackage{graphicx}
\usepackage{textcomp}
\usepackage{xcolor}
\usepackage{booktabs}
\usepackage{url}
\usepackage{hyperref}
\hypersetup{colorlinks=true,citecolor=blue,urlcolor=blue,linkcolor=red}

\begin{document}

\title{SwarmDrive: Semantic V2V Coordination for Latency-Constrained Cooperative Autonomous Driving}

\author{\IEEEauthorblockN{Anjie Qiu, Donglin Wang, Zexin Fang, Sanket Partani and Hans D. Schotten}
\IEEEauthorblockA{\textit{Institute for Wireless Communication and Navigation} \\
\textit{RPTU University Kaiserslautern-Landau}\\
Kaiserslautern, Germany \\
\{aqiu, dwang, zfang, partani, schotten\}@rptu.de}
}

\maketitle

\begin{abstract}
Cloud-hosted LLM inference for autonomous driving adds round-trip delay and depends on stable connectivity, while purely local edge models struggle under occlusion. We present \textit{SwarmDrive}, a semantic Vehicle-to-Vehicle (V2V) coordination framework in which nearby vehicles run local Small Language Models (SLMs), share compact intent distributions only when uncertainty is high, and fuse them through event-triggered consensus. We evaluate SwarmDrive in a 5-seed executable study built around one occluded-intersection case, combining matched operating-point comparisons with robustness sweeps. In that setting, SwarmDrive under its 6G communication setting (``Swarm 6G'') raises success from 68.9\% to 94.1\% over a single local SLM while reducing latency from a 510~ms cloud reference to 151.4~ms. However, increased number of participating vehicles shows increasedcommunication overhead and packet loss. Our SwarmDrive also evaluates the impact of swarm-size, packet-loss, and entropy-threshold sweeps and shows that the cooperative gain holds across ablations and is best balanced near an active swarm size of $N=4$ vehicles and an entropy trigger threshold of $\tau \approx 0.65$ in the current prototype. These results show that semantic edge cooperation can work under tight latency constraints in the targeted intersection case, but they are not a deployment-grade validation of a real 6G stack.
\end{abstract}


\section{Introduction}
Autonomous driving has long relied on modular perception-planning-control stacks \cite{yurtsever2020survey}, but long-tail traffic scenes still expose the limits of rule-centric pipelines. Recent language-model-based driving agents suggest that semantic reasoning can improve scene interpretation and decision support \cite{cui2023drive, mao2023gptdriver, xu2023driveable, cui2024llm4ad, wu2025multiagentllmsurvey, cui2025cotsurvey}. The difficulty is that many of these systems depend on large multimodal models or remote inference paths that do not fit onboard latency and compute budgets \cite{wang2023drivemlm, chen2024driving, wen2023dilu}. For the urban intersection case studied here, reaction windows are roughly 150--250~ms \cite{wang2023drivemlm, xu2023driveable}, so cloud round trips are costly and add reliability and privacy concerns \cite{chen2024driving}.

That pushes the problem back to the edge, where smaller local models are easier to deploy but more exposed to occlusion and partial observability. Our question is whether a cooperative swarm of edge-deployed SLMs can recover missing context through compact intent sharing. Drawing on cooperative vehicular perception, communication-aware collaboration \cite{xu2022opv2v, xu2022v2xvit, hu2022where2comm}, and recent vehicle-to-vehicle LLM coordination \cite{chiu2025v2vllm}, \textit{SwarmDrive} exchanges post-inference intent distributions instead of raw sensor streams. The payloads stay lightweight, but neighboring vehicles can still contribute complementary viewpoints when the scene is ambiguous, consistent with connected-driving and AI-enabled V2X studies that emphasize compact cooperation under communication constraints \cite{partani2021connecteddriving, wang2025survey6gv2x}.

This paper makes four contributions. We define the \textit{SwarmDrive} architecture for decentralized scene understanding and event-triggered consensus among edge-deployed SLM nodes. We build a reproducible prototype workflow with a DiLu-style semantic interface~\cite{wen2023dilu}, report an intersection-centered comparison of local, baseline V2X, and Swarm 6G operating points, and study robustness through swarm-size, packet-loss, and entropy-threshold analyses.

\begin{figure*}[!t]
\centering
\includegraphics[width=\textwidth]{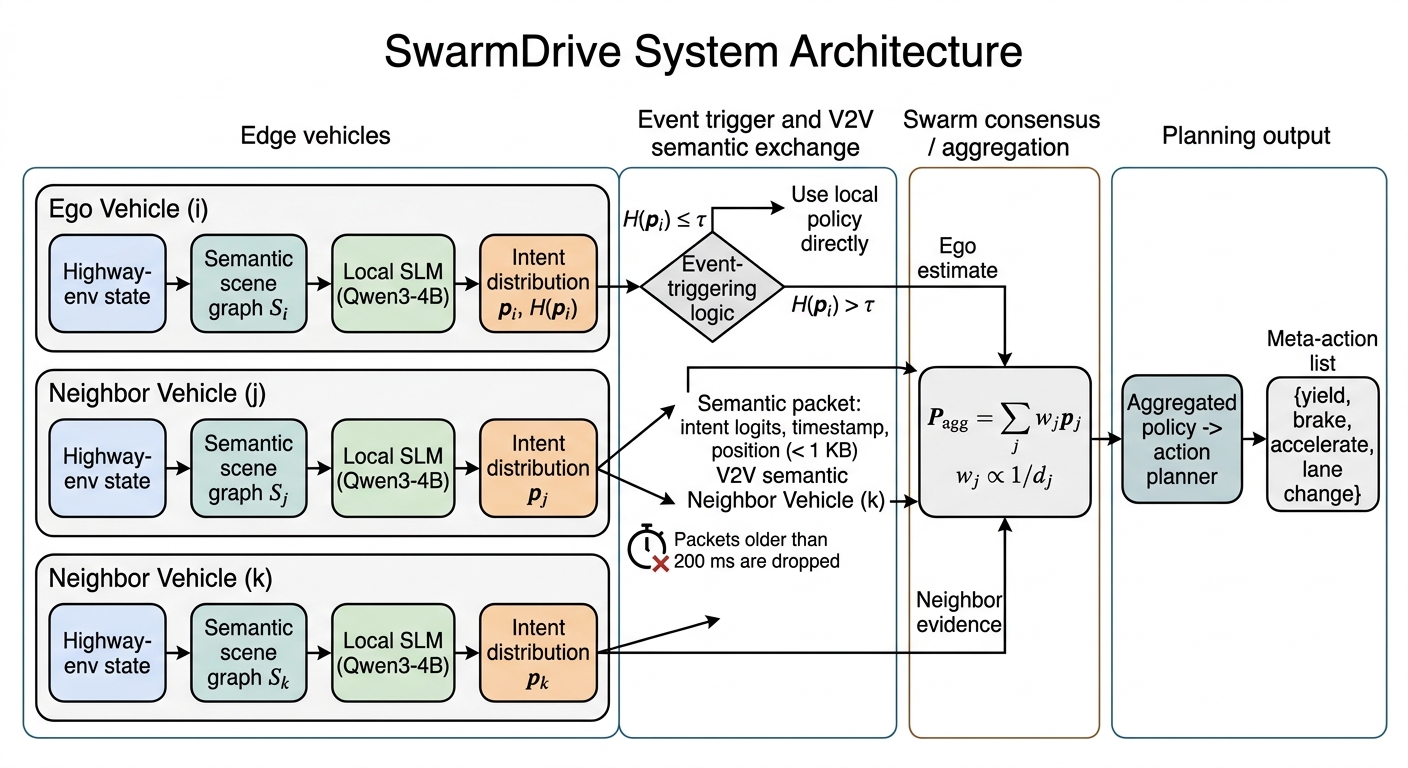}
\caption{System architecture of SwarmDrive. Individual vehicles process multimodal perception inputs locally using edge-deployable SLMs. Instead of forwarding raw data, vehicles share compact intent-level messages over V2V links, and the Swarm Consensus Module aggregates them into a driving policy.}
\label{fig:method_architecture}
\end{figure*}

\section{Related Work}

\subsection{Language Models and Edge Intelligence for Autonomous Driving}
Recent work has moved language-model-based autonomous driving from descriptive assistants toward closed-loop decision support, showing that semantic reasoning can improve scene interpretation and action selection in complex traffic scenarios \cite{cui2024llm4ad, xu2023driveable}. The main systems challenge is practical: many LLM/VLM driving pipelines still rely on large models or remote inference paths that do not fit vehicular latency and compute budgets \cite{wang2023drivemlm, chen2024driving}. This is why edge-side deployments built around smaller models and compact semantic interfaces matter; Qwen3-4B is a more realistic SLM scale for on-vehicle experiments, but single-vehicle local reasoning still struggles under occlusion and partial observability \cite{wang2025survey6gv2x, qwen2025qwen3}.

\subsection{Semantic V2V Coordination and Communication-Aware Cooperation}
Multi-vehicle cooperation addresses that blind-spot problem by letting nearby agents share complementary evidence under uncertainty. Recent V2V-LLM work studies multimodal language-model coordination across neighboring vehicles \cite{chiu2025v2vllm}, while communication research shows that retransmission design, numerology, and channel congestion directly shape cooperative performance \cite{wang2020retransmissions, wang2023numerology, qiu2020broadcaststorm}. Unlike V2V-LLM, which fuses multimodal features across a shared LLM backbone, SwarmDrive exchanges post-inference intent distributions. That decouples the model choice on each vehicle and keeps message size in the sub-kilobyte range.

A closely related line of work is \textit{cooperative perception}, where vehicles share compressed intermediate representations over V2V links to improve downstream perception \cite{xu2022opv2v, hu2022where2comm}. SwarmDrive sits one layer higher: instead of fusing feature tensors, it aggregates intent distributions produced by local language-model reasoning. The two lines of work are complementary, and cooperative perception could serve as an upstream source of richer semantic context.

\section{Methodology}\label{sec:method}

SwarmDrive uses a hierarchical decentralized architecture in which edge-capable nodes exchange intent only when needed. This section outlines the system design and the prototype data-collection process.

\subsection{System Architecture}
\begin{figure*}[!t]
\centering
\includegraphics[width=0.9\textwidth]{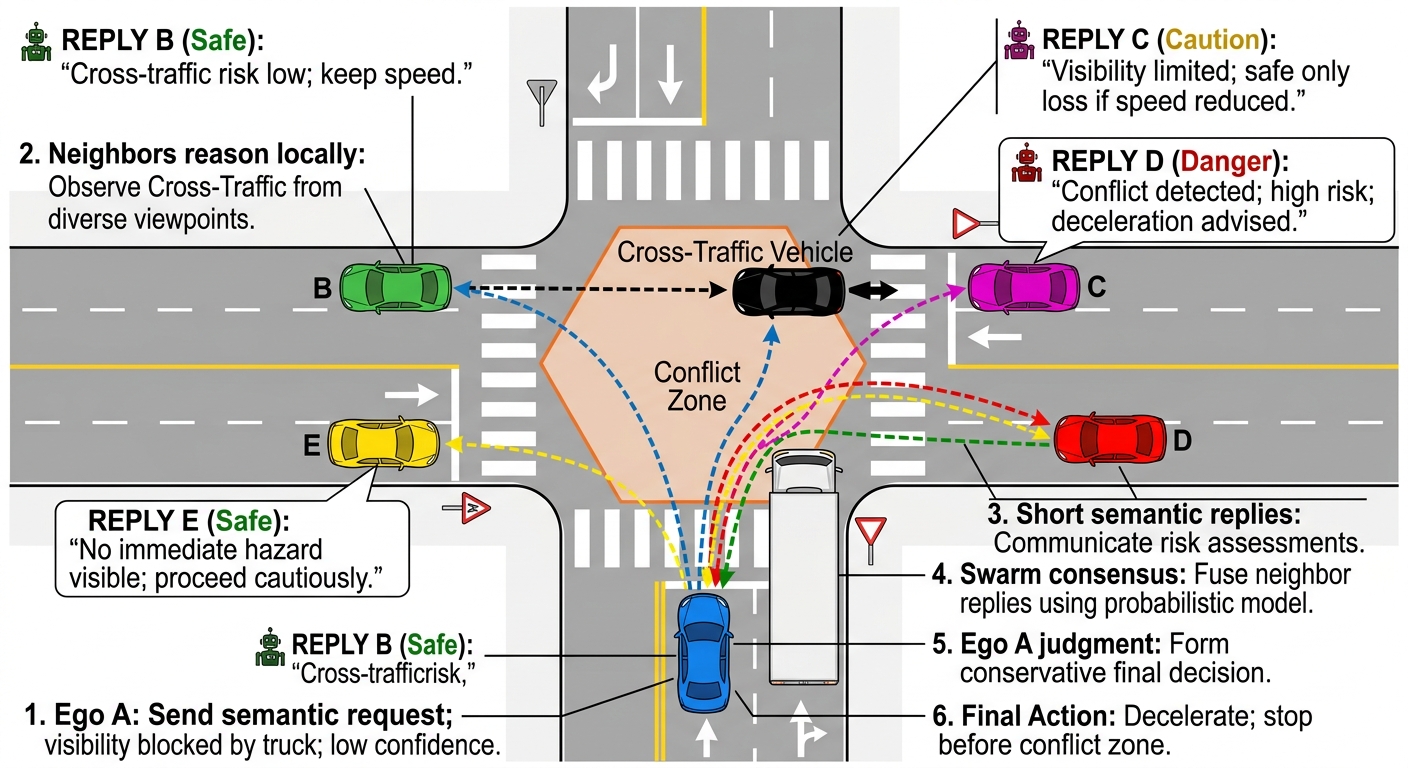}
\caption{Semantic messaging example for the canonical occluded-intersection scenario. Ego~A requests cooperative support under structural occlusion, neighboring vehicles provide viewpoint-dependent risk cues, and the fused response leads to a conservative action before the conflict zone.}
\label{fig:scenario}
\end{figure*}

As shown in Fig.~\ref{fig:method_architecture}, SwarmDrive runs without cloud intervention and places three modules on each vehicle's edge-computing unit: a Multimodal Perception Module, a Local SLM Inference Engine, and a Swarm Consensus Module. In the current \texttt{highway-env} prototype, the interface follows a DiLu-style semantic loop~\cite{wen2023dilu} in which synchronized LiDAR, radar, and camera observations are converted into an ego-centric scene graph $\mathcal{S}_i$ rather than exchanged as raw tensors (see Fig.~\ref{fig:scenario}).

The \textit{Local SLM} maps $\mathcal{S}_i$ to a distribution over meta-actions $A = \{a_1, a_2, \dots, a_k\}$, yielding the local intent estimate $\mathbf{p}_i = M(\mathcal{S}_i)$. When the estimate is confident, the vehicle stays in local mode; otherwise it requests additional intent-level evidence from neighboring vehicles.

When the local estimate is uncertain, the \textit{Swarm Consensus Module} initiates a lightweight communication step. Rather than transmitting raw sensor payloads, vehicle $i$ broadcasts only its intent distribution $\mathbf{p}_i$ and its spatial coordinates over DSRC or C-V2X links to neighboring vehicles within radius $R$. After collecting intent distributions from the relevant neighbors, the consensus module computes an aggregated policy $\mathbf{P}_{agg}$ using
\begin{equation}
\mathbf{P}_{agg} = \sum_{j=1}^{N} w_j \mathbf{p}_j
\end{equation}
The weights satisfy the normalization constraint $\sum_{j=1}^{N} w_j = 1$, and are explicitly defined as:
\begin{equation}
w_j = \frac{1/d_j}{\sum_{k=1}^{N} 1/d_k}
\end{equation}
where $d_j$ denotes the Euclidean distance from vehicle $j$ to the centroid of the conflict zone. This inverse-distance weighting prioritizes viewpoints closest to the potential hazard and reduces the impact of single-view occlusion errors. Neighbor intent vectors arrive asynchronously, so only vectors within a nominal 200~ms decision window are aggregated; otherwise the ego vehicle falls back to $\mathbf{p}_i$.

\subsection{Data Collection and Consensus Triggering}
SwarmDrive is simulator-agnostic, but the current evidence comes from an executable prototype that keeps the same semantic scene-graph, trigger, consensus, and channel modules across both a \texttt{highway-env}-aligned scenario interface and a deterministic proxy backend for larger campaign sweeps. This follows the broader use of lightweight OpenAI-Gym-style platforms for rapid ML and V2X prototyping \cite{qiu2023openaigym}. In each run, local observations are serialized into compact scene graphs, the SLM produces an intent distribution, and the vehicle queries neighbors only when uncertainty is high. To avoid constant high-rate broadcasting, consensus is event-triggered:
\begin{equation}
H(\mathbf{p}_i) = -\sum_{k} p_{i,k} \log p_{i,k} > \tau
\end{equation}
When the entropy condition is not met, the vehicle stays in local mode; otherwise it broadcasts its intent vector and fuses valid neighbor responses. The same branch-level path is reused later for the multi-seed robustness sweeps, where the proxy backend makes controlled ablations tractable while keeping the measured quantities tied to the executable branch rather than to a deployment-grade stack.

\subsection{Prompt Engineering for Multimodal Perception}
\label{sec:prompt}
In the prototype, heterogeneous observations are compressed into a bounded multimodal prompt instead of being passed as raw tensors. Camera frames enter as Qwen-compatible image tokens, while LiDAR and radar detections are serialized into compact scene-graph records containing object type, relative geometry, and kinematics. Rather than using one generic instruction for every node, the prompt design separates two roles: neighboring vehicles produce compact cooperative messages from their own viewpoints, while the ego vehicle merges those messages with its local scene graph and returns the final action distribution over the meta-action set $A$.
\begin{quote}\small
\textbf{Helper-vehicle prompt:} ``You are a neighboring vehicle assisting the ego car. From your local view, report only the most decision-relevant evidence for the conflict zone, summarize the visible risk, and send one compact semantic message with a recommended maneuver and confidence score.''
\end{quote}
\begin{quote}\small
\textbf{Ego-vehicle prompt:} ``You are the ego vehicle. Combine your local scene graph with the received cooperative messages, reason step by step over occlusion, conflict-zone risk, and message agreement, and then output only the final JSON action distribution over $A$. Prioritize safety over efficiency when the evidence is uncertain or conflicting.''
\end{quote}
This split keeps the scene description within a predictable 200--400-token budget, preserves lightweight neighbor messages, and makes the message-fusion stage explicit while still supporting reproducible parsing of the final output.

\section{Experimental Setup}

\subsection{Evaluation Scenarios}\label{sec:scenario}
The paper focuses on the occluded-intersection scenario in Fig.~\ref{fig:scenario} because it isolates the structural-occlusion failure mode that motivates SwarmDrive. In the executable campaign, the \texttt{intersection} preset uses an ego speed of 14~m/s, a conflict-zone distance proxy of 3.5~m, and a cooperative neighborhood of three nearby vehicles. Fig.~\ref{fig:scenario} should be read as the anchor scene for the paper's main claim rather than as one member of a broader scenario family.

\subsection{Baselines and Communication Profiles}
All settings share the same semantic scene-graph abstraction, local action vocabulary, and planner interface; only the communication path changes. The paper compares three matched operating points within the canonical intersection case: \textit{single local} disables V2V, \textit{swarm baseline V2X} uses a conventional proxy V2X profile, and \textit{swarm 6G} denotes the paper's 6G communication setting rather than a standards-compliant 6G PHY/MAC stack. In the executable branch, the baseline channel uses 25~ms nominal latency, 8\% packet loss, and 10~Mbps bandwidth, whereas the 6G setting uses 5~ms latency, 1.2\% loss, 200~Mbps bandwidth, and optional roadside-edge assistance. These are not standard-compliant link budgets; they are calibrated assumptions chosen to remain broadly consistent with prior C-V2X retransmission/numerology studies and VANET relay-reliability analyses \cite{wang2019simulators, wang2020retransmissions, wang2023numerology, qiu2020broadcaststorm}. For the swarm-size sweep in Fig.~\ref{fig:results_mechanism}, the channel further increases effective packet loss with active swarm size to reflect contention from denser event-triggered intent exchanges.

\subsection{Protocol and Metrics}
The reported study uses deterministic prototype defaults with fixed seeds for environment reset, communication noise, and local policy noise. The primary operating-point comparison uses 5 seeds and 1,500 episodes per condition, whereas sweep-style ablations use 5 seeds and 300 episodes per grid point. The entropy trigger threshold is fixed at $\tau = 0.7$ for the matched operating-point comparisons, based on a separate threshold sweep reported later in Fig.~\ref{fig:results_mechanism}. Each run exports both per-condition summary rows and per-episode records as stable CSV/JSONL artifacts, so the same executable branch supports both figure generation and later audit of the underlying numbers.

The paper reports implementation-aligned metrics: \texttt{latency\_proxy} is the mean decision latency across the control cycle, \texttt{trigger\_rate} the mean fraction of control steps in which consensus is triggered, \texttt{message\_count} the total number of V2V transmissions, and \texttt{success\_rate} the fraction of episodes completed without collision or forced failure. For readability, the communication-overhead panel later reports mean messages per episode, while the raw exports retain total \texttt{message\_count}. We also report an effective V2V packet-loss ratio under the configured or density-coupled channel model. Unless noted otherwise, error bars denote 95\% confidence intervals across seed means. A Cloud GPT-4 operating point appears only as a paper-level summary comparator in Fig.~\ref{fig:results_core}; it is not a runnable baseline in the executable branch and is therefore shown without error bars.

\section{Results}

Unless noted otherwise, the results in this section come from the calibrated executable campaign for the occluded-intersection case, whereas the 510~ms cloud point remains a paper-level comparator only. Throughout this section, `Swarm 6G` is used as a readable shorthand for the manuscript's 6G communication setting.

\subsection{Composite Figure A: Core Operating Points}
\begin{figure*}[!t]
\centering
\includegraphics[width=\textwidth]{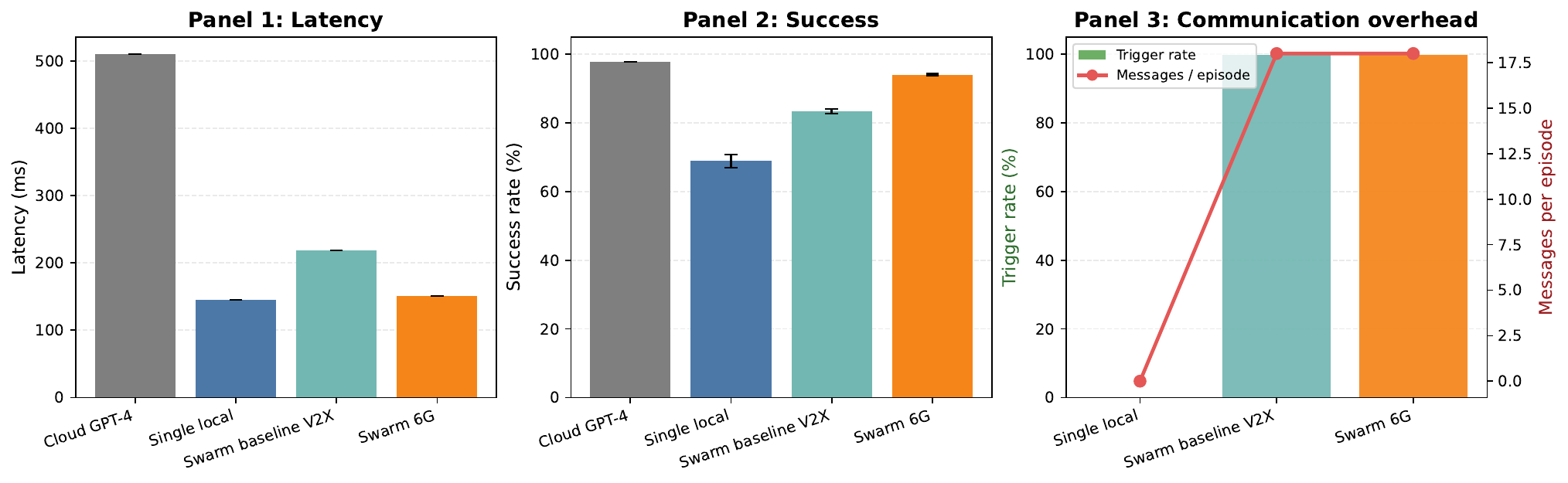}
\caption{Composite Figure A: matched operating-point comparison for the canonical occluded-intersection case, showing latency, success rate, and executable-only communication overhead for the Cloud GPT-4 paper-level reference and the three executable conditions. Executable conditions use 5 seeds and 1,500 episodes per seed; error bars denote 95\% confidence intervals across seed means.}
\label{fig:results_core}
\end{figure*}

Fig.~\ref{fig:results_core} summarizes the main operating-point comparison for the occluded-intersection case. The Cloud GPT-4 reference is the slowest point at 510~ms per decision cycle, well outside the practical reaction window of the target scenario. Among the executable conditions, the single-local SLM is the latency minimum at 145.0~ms, but it succeeds in only 68.9\% of episodes because the ego vehicle has no access to complementary viewpoints under structural occlusion.

\begin{figure*}[!t]
\centering
\includegraphics[width=\textwidth]{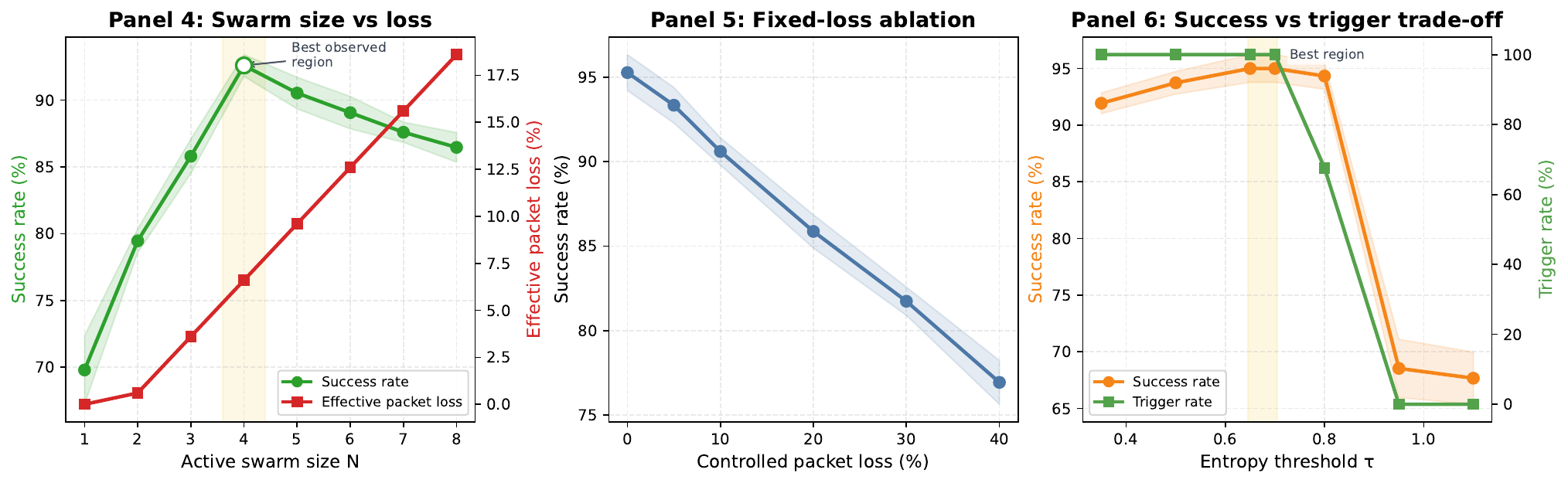}
\caption{Composite Figure B: mechanism and robustness analyses for swarm size, fixed packet loss, and entropy-trigger threshold $\tau$ in the canonical occluded-intersection case. All panels use 5 seeds and 300 episodes per grid point; error bands denote 95\% confidence intervals across seed means.}
\label{fig:results_mechanism}
\end{figure*}

The baseline V2X swarm recovers much of that lost capability, raising success to 83.5\%, but it does so at 218.9~ms because this intersection induces a nearly always-triggered cooperation pattern and the baseline V2X profile pays the full per-message latency budget. The Swarm 6G profile, i.e., our 6G communication setting, follows the same operating pattern, still averaging 18 messages per episode and a 100\% step-level trigger rate in this highly ambiguous setting, but reaches 94.1\% success at 151.4~ms because the low-latency, low-loss 6G assumption sharply reduces per-message latency and effective loss. Panel 3 makes that asymmetry explicit by excluding the cloud comparator from the communication subplot while keeping the trigger-rate versus message-volume legend inside the panel. The result is straightforward: this 6G setting recovers most of the cloud-reference success while staying much closer to the local-latency regime than to the baseline V2X operating point. That supports a feasibility claim under calibrated communication assumptions, not a validation of a deployment-grade 6G stack.

\subsection{Composite Figure B: Mechanism and Robustness}\label{sec:sensitivity}
Fig.~\ref{fig:results_mechanism} helps explain where the cooperative gain comes from. Panel 4 is intentionally different from Fig.~\ref{fig:results_core}: it uses a density-coupled loss model rather than the fixed canonical operating point. Under that harsher sweep, success rises from 69.8\% at $N=1$ to 79.5\% at $N=2$, 85.8\% at $N=3$, and 92.6\% at $N=4$, after which contention begins to dominate. The same sweep shows effective packet loss climbing from 0.6\% at $N=2$ to 6.6\% at $N=4$ and 18.6\% at $N=8$, with success easing back to 86.5\% at the largest swarm. The best region is around $N=3$--4: large enough to recover multi-view context, but not yet so dense that density-coupled degradation erodes the gain. Importantly, the $N=4$ peak remains only about 1.5 percentage points below the fixed-profile 6G core point in Fig.~\ref{fig:results_core}, so the two panels remain consistent.

Panel 5 isolates packet loss directly and shows gradual rather than catastrophic degradation. Success is 95.3\% at 0\% loss, 90.6\% at 10\% loss, 85.9\% at 20\% loss, and 76.9\% at 40\% loss. Because SwarmDrive uses event-triggered consensus instead of a permanently synchronized swarm, dropped messages degrade the fused estimate before they fully collapse the control loop; the ego node can still fall back to local reasoning when support is missing. The controlled-loss curve is also consistent with Panel 4: the 6.6\% effective loss observed at $N=4$ maps to a low-90\% success regime, while the 18.6\% loss at $N=8$ maps to the mid-to-high-80\% range seen in the density-coupled sweep.

Panel 6 sweeps the entropy threshold $\tau$ and places the best operating region around $\tau = 0.65$--0.70, where success reaches 95.0\%. The highlighted band in the figure is descriptive, not inferential; it marks the region supported by the sweep. Lower thresholds over-trigger communication and reduce success to 91.9--93.7\% even though the trigger rate remains saturated at 100\% in this highly ambiguous intersection. Raising the threshold to $\tau = 0.8$ still preserves most of the success (94.3\%) while dropping the step-level trigger rate to 67.6\%, showing that entropy gating does affect communication behavior in the executable branch. Very high thresholds suppress cooperation entirely: once $\tau \ge 0.95$, the trigger rate drops to 0 and success falls back to roughly 67.7--68.5\%, effectively the disconnected local regime. The sweep also gives a concrete mechanism-level reason for fixing $\tau = 0.7$ in the matched operating-point comparison.

\subsection{Limitations}
The current results suite is more structured than a single-point comparison, but it is still a prototype study rather than a homologation-grade benchmark. The operating-point and mechanism panels come from a calibrated executable campaign whose exported manifest records the proxy-model version and calibration label; that keeps the figure values reproducible, but it is still not a full closed-loop \texttt{highway-env} or standards-compliant PHY/MAC co-simulation for every sweep. The density-coupled packet-loss curve in Fig.~\ref{fig:results_mechanism} is a controlled contention assumption rather than the output of a full communication stack. The Cloud GPT-4 point remains a paper-level comparator rather than a runnable baseline. The communication panels show how event triggering behaves within the targeted occluded-intersection case, but the present campaign still does not directly validate bandwidth or energy savings on real hardware, so the paper supports semantic 6G edge advantage in a calibrated executable campaign rather than deployment-grade 6G communication claims.

\section{Conclusion}
This paper presented SwarmDrive as a decentralized architecture for uncertainty-triggered cooperative decision support in autonomous driving. The evaluation is intentionally centered on one occluded-intersection case, where the main claim can be tested directly against the structural-occlusion failure mode that motivates the method. In that setting, SwarmDrive under its 6G communication setting improves success from 68.9\% to 94.1\% over a single local SLM while reducing latency from the 510~ms cloud-reference operating point to 151.4~ms. The mechanism sweeps show that the same cooperative advantage remains clear under swarm-size, packet-loss, and entropy-threshold variation within that targeted setting. Taken together, the results support semantic low-latency 6G edge cooperation in the intersection case, not direct real-world 6G performance.

Future work will close the remaining realism gap by re-running the intersection study in richer \texttt{highway-env} or hardware-in-the-loop settings, adding roadside-edge assistance and richer channel realism, comparing adaptive communication scheduling against fixed event triggering, and validating trust- and fault-aware consensus as well as communication and energy budgets on real edge platforms. Extending the same semantic cooperation pipeline to other scenario families such as \texttt{highway}, \texttt{merge}, and \texttt{parking} is left explicitly out of scope for the present paper.

\section*{Acknowledgments}
This work has been supported by the Bundesministerium für ­Forschung, Technologie und Raumfahrt (BMFTR) as part of the Open6G+ project with FKZ 16KIS2406. The authors alone are responsible for the content of the paper.

\bibliographystyle{IEEEtran}
\bibliography{references}

\end{document}